\documentclass[11pt]{article}

\usepackage{acl}

\usepackage{xcolor}
\usepackage{subcaption}
\usepackage{rotating}
\usepackage{amsmath}
\usepackage{bm}
\usepackage{multirow}
\usepackage{xspace}
\usepackage{latexsym}
\usepackage{graphicx}
\graphicspath{ {./imgs/} }
\usepackage{hyperref}
\usepackage{array}
\usepackage{float}
\usepackage{booktabs}
\usepackage{color, colortbl}
\usepackage[inline]{enumitem}
\usepackage[flushleft]{threeparttable}

\usepackage{times}
\usepackage[T1]{fontenc}

\usepackage[utf8]{inputenc}

\usepackage{microtype}

\long\def\eat#1{}

\def\dataset{\textsc{XtremeSpeech}\xspace}
\def\dictionary{\textsc{HateLexicon}\xspace}
\def\lime{{\fontfamily{lmr}\selectfont LIME}\xspace}

\title{Sociocultural knowledge is needed for selection of shots in hate speech detection tasks}

\author{ 
	Antonis Maronikolakis$^{1,2}$ \quad 
	Abdullatif Köksal$^{1,2}$ \quad
	Hinrich Schütze$^{1,2}$ \quad \\ 
	$^1$Center for Information and Language Processing, LMU Munich \\
	$^2$Munich Center for Machine Learning \\ 
	{\tt antmarakis@cis.lmu.de akoksal@cis.lmu.de}
}

\def\figref#1{Figure~\ref{fig:#1}}
\def\figlabel#1{\label{fig:#1}\label{p:#1}}

\def\tabref#1{Table~\ref{tab:#1}}

\def\tablabel#1{\label{tab:#1}\label{p:#1}}

\def\seclabel#1{\label{sec:#1}}

\newcounter{notecounter}
\newcommand{\enotesoff}{\long\gdef\enote##1##2{}}
\newcommand{\enoteson}{\long\gdef\enote##1##2{{
		\stepcounter{notecounter}
		{\large\textbf{ \hspace{1cm}\arabic{notecounter} $<<<$ ##1: ##2 $>>>$\hspace{1cm}}}}}}
\enoteson
\enotesoff                      

\definecolor{myblue}{rgb}{0,0,.5}
\definecolor{myred}{rgb}{1,0,0}
\definecolor{mypurple}{rgb}{.5,0,.5}

\begin{document}
\maketitle
\begin{abstract}
	We introduce \dictionary, a lexicon of slurs and targets of hate speech for Brazil, Germany, India and Kenya, to aid model development and interpretability.
	First, we demonstrate how \dictionary can be used to interpret model predictions, showing that models developed to classify extreme speech rely heavily on target words.
	Further, we propose a culturally-informed method to aid shot selection for training in low-resource settings. In few-shot learning, selection of shots is of paramount importance to model performance and we need to ensure we make the most of available data.
	We work with HASOC German and Hindi data for training and the Multilingual HateCheck (MHC) benchmark for evaluation. We show that selecting shots based on our lexicon leads to models performing better than models trained on shots sampled randomly. Thus, when given only a few training examples, using \dictionary to select shots containing more sociocultural information leads to better few-shot performance. With these two use-cases we show how our \dictionary can be used for more effective hate speech detection.
\end{abstract}

\section{Introduction}
\seclabel{intro}

To curb the spread and dissemination of hate speech online, the research and industry communities have focused on the collection of hate speech data from social media and the development of models to automatically filter out harmful content.

While there have been efforts to cover multiple languages
\cite{multi_hate_speech,hasoc,refugee_crisis_german,xtremespeech},
most work is still conducted for English settings
\cite{davidson,founta,social_bias_frames}.
Concurrently, it has been shown that cross-lingual transfer capabilities of models are limited in this domain \cite{hatespeech_zero_shot_cross_lingual,hatespeech_crosslingual_embds}, potentially due to the heavily culture-specific and subjective nature of the data. Thus, leveraging high-resource languages to aid performance in low-resource ones is not a
reliable
option. Instead, methods need to be developed to better utilize the available data.

Towards efforts for more inclusive hate speech research, we are introducing \dictionary, a lexicon of slurs and target group denotations that can be used as an aid to model training and interpretability. Curated in collaboration with members familiar with sociopolitical balances in the examined countries (Brazil, Germany, India and Kenya), \dictionary aims to bring cultural knowledge to hate speech model development.

\begin{table}[]
	\centering
	\begin{tabular}{lcc}
		\toprule
		Set & German & Hindi \\ \midrule
		Random$_{64}$ & 0.51$_{3.6}$  & 0.47$_{2.5}$ \\
		Random$_{96}$ & 0.53$_{4.7}$  & 0.46$_{3.2}$ \\ \midrule
		Lexicon$_{64}$ & 0.54$_{1.8}$ & 0.50$_{5.4}$ \\
		Lexicon$_{96}$ & \textbf{0.55$_{\bm{1.0}}$} & \textbf{0.52$_{\bm{1.1}}$} \\ \midrule
		All$_{128}$ & 0.53$_{2.1}$ & 0.44$_{5.6}$ \\ \bottomrule
	\end{tabular}
	\caption{Comparison of F1-scores for German and Hindi between randomly- and \dictionary-sampled training sets of sizes 64 and 96. Standard deviation in subscript.}
	\tablabel{pitch}
\end{table}

\textbf{Models often rely on keywords for predictions.} While this can be an effective
tactic in developing baselines (e.g.,
keyword-based models), it can have undesirable effects,
such as associating generally innocuous terms with extreme
speech; e.g.,
a negative interpretation of
the term `Muslim'.
\textbf{This erroneous association between target group names and hate speech may lead to further marginalization of vulnerable groups}, misclassifying text mentioning these terms with hate speech and consequently filtering them out.
Further, seemingly \textbf{innocuous terms have been appropriated by extreme speech peddlers and may not be picked up by models}, such as the term `Goldstücke' in German (originally meaning `gold pieces' and appropriated to refer to refugees in a derogatory manner). Models unable to recognize these keywords as hateful in certain contexts will lead to hate speech falling through the cracks.

This is especially salient in few-shot settings, where the wide range of targets and slurs might not be adequately captured in annotated datasets. Further, it has been shown that model performance fluctuates a lot depending on the selection of training shots \cite{zheng-etal-2022-fewnlu}.
Therefore, \textbf{we need a better strategy to make the most out of the available data to select shots more conducive to model performance}.

Motivated by the above problems, we propose \dictionary, a lexicon aiming to  \begin{enumerate*}[label=(\roman*)]
	\item aid model interpretability by providing ground-truth labels on common terms and
	\item improve shot selection in low-resource settings by better coverage of key terms such as targets and slurs
\end{enumerate*}. To create \dictionary, we collaborated with annotators from our examined countries, who provided a list of
keywords and marked them as target groups, slurs, neutral words or any combination of these labels.

Our contributions, in short, are the following:

\begin{enumerate}
	\item We introduce \dictionary, a lexicon of target group names and slurs from Brazil, Germany, India and Kenya.
	\item We show how cultural information can aid in model interpretability, identifying which slurs and targets affect performance the most.
	\item We show that culturally-informed sampling outperforms random sampling in few-shot hate speech detection settings (\tabref{pitch}).
	\item We propose a method to complement existing datasets by querying data using \dictionary terms.
\end{enumerate}

\section{Related Work}

\textbf{Hate Speech Detection.} Nascent efforts to tackle hate speech focused on the curation of general-purpose, English datasets \cite{founta,davidson}, later expanding into more granular annotation \cite{misogyny_annotated_data,hate_towards_political_opponent,refugee_crisis_german,social_bias_frames,incivility_in_news,implicitly_abusive_language}. Most work is performed on datasets of thousands of examples, allowing for straightforward finetuning of models. In our work, we focus on low-resource settings where only a few examples are available for training and thus traditional finetuning techniques cannot be applied.

Recently, work has been conducted to cover a larger range of languages \cite{multi_hate_speech,hatespeech_crosslingual_embds,xtremespeech,plakidis-rehm-2022-dataset}. In our work, we continue previous efforts into multilingual hate speech detection by proposing a lexicon of terms (pertinent to the domain hate speech) for Brazilian Portuguese, English, German, Hindi and Swahili.

Analysis has taken place both on the model and the dataset level \cite{hatexplain,hatespeech_biased_datasets,hate_speech_data_analysis,intersectional_bias,hate_speech_cross_dataset,racial_bias_in_data}.
Further, hate speech datasets have been examined for presence and reproduction of bias \cite{racial_bias_in_hatespeech,civil_text_rephrasing,aae_bias_hatespeech,maronikolakis-etal-2022-analyzing}.
We continue in this direction by proposing a lexicon that can aid in interpretability and model analysis efforts.

Previous work has uncovered annotator bias \cite{refugee_crisis_german,are_you_racist,global_workforce_demographics,ethical_implications_crowdsourcing,annotator_bias_demographic}, with work conducted to propose frameworks of ethical data curation \cite{ethical_scaling,collecting_sociocultural_data,convenience_or_death,hate_speech_frontiers,gebru_gender_race,decolonial_ai_theory}. To mitigate bias in our work, we are directly working with community-embedded members.

\citet{hatecheck} proposed a benchmark for unified evaluation of hate speech detection models in English, subsequently expanded into the Multilingual HateCheck (MHC) benchmark for multiple languages \cite{rottger-etal-2022-multilingual}, used in our work.

\textbf{Few-shot Learning.} Large language models exhibit zero- and few-shot capabilities \cite{gpt3,wei2022emergent,sanh2022multitask,le-scao-rush-2021-many,gao-etal-2021-making,schick-schutze-2021-just}.
A challenge with finetuning large language models (and few-shot learning in particular) is inconsistency: the selection of training data greatly affects performance \cite{zheng-etal-2022-fewnlu,mosbach2021on,lu-etal-2022-fantastically}.
Since in few-shot learning settings only a few examples are available, any noise in the data can exacerbate training issues \cite{köksal2022meal}. In our work we propose a lexicon-based approach to shot selection that consistently improves performance.

Earlier work in few-shot learning focused on prompt-based training \citet{schick-schutze-2021-exploiting,https://doi.org/10.48550/arxiv.2204.14264,shin-etal-2020-autoprompt,logan-iv-etal-2022-cutting,zhao-schutze-2021-discrete}. \citet{setfit} introduced a prompt-free approach to learning from small datasets (SetFit). Through the use of SentenceBERT and its Siamese-network training paradigm \cite{sentencebert}, SetFit 
generates pairs of training examples and learns to minimize the distance of training example representations
of the same class and, conversely, to maximize the distance for examples from different classes. This process results in 
a model that can generate strong sentence embeddings, which can be then used to train a classification head on a task. In our work, we use SetFit to train a multilingual SentenceBERT model on German and Hindi.

\section{Methodology}

To showcase the usefulness of \dictionary in hate speech model development, we examine two use cases:
model interpretability and few-shot model development, showcasing how \dictionary can be utilized to improve both processes in the hate speech domain.

\subsection{\dictionary Curation}

For the curation of \dictionary, we employed\footnote{All annotators were paid the same rate, which was above minimum wage in all countries.} annotators to provide slurs, target group names and neutral words that appear often in hateful contexts online. We employed three annotators in Brazil, four in Germany, four in India and two in Kenya.

The annotators were tasked with providing terms alongside a short description. The sourcing of terms was left up to the annotators. We suggested they could use social media (e.g., searching for certain hateful hashtags or groups), but no restrictions were imposed. Instead, we relied on the sociocultural knowledge of the annotators to guide curation. We allowed for coordination between the annotators, but with no explicit instructions to actively collaborate.
Terms are written in Brazilian Portuguese, English, German, Hindi or Swahili. Acceptable terms are: \begin{enumerate*}[label=(\roman*)]
	\item slurs attacking the identity of a person or group, such as ethnicity, religion and sexuality.
	\item target group denotations, such as religious groups (e.g., `Muslim') and marginalized communities (e.g., `homosexual').
	\item neutral words that may appear often in hateful contexts or datasets (e.g., `Frauenquote', in German meaning `quota of/for women')
\end{enumerate*}. Statistics and indicative entries are shown in \tabref{lexicon_stats} and \tabref{lexicon_examples}.

To evaluate the quality of our lexicon, terms submitted by one annotator were cross-checked by the other annotators of the same country.
From discussions with annotator teams, it was made clear that a few terms can be assigned more than one type. For example, in German, the term `Schwule' can be used by \textit{homosexuals to describe themselves or as a slur against them}. In these instances, we allow annotation with multiple types. For example, `Schwule' is annotated both as a target group denotation \textit{and} a slur, to better capture the dual nature of the word.

\begin{table}[]
	\centering
	\small
	\begin{tabular}{ccccc}
		\toprule
		Type & Brazil & Germany & India & Kenya \\ \midrule
		Neutral & 30 & 4 & 3 & 21 \\
		Target & 4 & 3 & 7 & 29 \\
		Slur & 11 & 18 & 35 & 43 \\
		Neutral/Target & 0 & 1 & 0 & 2 \\
		Neutral/Slur & 0 & 18 & 1 & 6 \\
		Target/Slur & 0 & 5 & 0 & 12 \\ 
		\textit{Total} & \textit{45} & \textit{50} & \textit{50} & \textit{116} \\ \bottomrule
	\end{tabular}
	\caption{\dictionary statistics for terms.}
	\tablabel{lexicon_stats}
\end{table}

\begin{table}[]
	\centering
	\small
	\renewcommand{\arraystretch}{1.25}
	\begin{tabular}{l|ccp{2cm}}
		\toprule
		\textbf{Country} & \textbf{Text} & \textbf{Type} & \textbf{Description} \\ \midrule
		Brazil & gorda & Slur & overweight women \\
		Brazil & traveco & Slur & transsexual \\
		Brazil & hora & Neutral & meaning `hour' \\ \midrule
		Germany & Flüchtling & Target & refugee \\
		Germany & Schwuchteln & Slur & derogatory term for homosexual \\
		Germany & Roma & Target & ethnic group \\ \midrule
		India & Bhimte & Slur & caste-ist term \\ 
		India & Mullo & Slur & Muslim people \\ 
		India & peaceful & Slur & Muslim people \\ \midrule
		Kenya & wakalee & Target & Kalenjin ethnic group \\ 
		Kenya & nugu & Slur & generic slur \\
		Kenya & foreskin & Slur & derogatory against uncircumcised Luo \\ \bottomrule
	\end{tabular}
	\caption{Example entries of \dictionary.}
	\tablabel{lexicon_examples}
\end{table}

\subsection{Interpretability}

We propose the use of \dictionary as a tool to interpret model predictions. Popular interpretability toolkits such as \lime \cite{lime} indicate which words are most associated with predictions of particular classes. In hate speech contexts, words most important for making predictions are oftentimes target group denotations
or slurs. While slurs are a more obvious indicator of hateful language, target group denotations also naturally appear in hateful contexts and there is the danger of
overemphasizing their association
with hate speech. This correlation could lead to further marginalization of target groups, with content mentioning target group denotations being filtered out as hate speech. With \dictionary we can investigate keywords associated with model predictions from a more culturally-informed perspective to better verify whether the model has accrued bias against these groups.

We take as an example use case the work in \dataset \cite{xtremespeech}, where a novel dataset of hate speech is introduced for Brazil, Germany, India and Kenya. In \citet{xtremespeech}, the authors use \lime to interpret their developed \texttt{mBERT} models, identifying words contributing the most to predictions.
In our work we operate on two levels: First, using \dictionary, we investigate the list of top-contributing words and show that in all examined countries, models emphasize heavily on target groups and slurs. Further, we examine the change of model representations for targets and slurs of the Kenyan and Indian subsets before and after model finetuning.

\dataset is a hate speech dataset with social
media texts collected from multiple online platforms and messaging apps. Languages covered in the dataset are Brazilian Portuguese, German, Hindi and Swahili, as well as English (either on its own or in the form of code switching with the native language).

All text in \dataset is targeting one or more groups based on protected attributes (e.g., women or religious minorities), annotated for three levels of extremity: derogatory, exclusionary and dangerous extreme speech. Brief descriptions (as defined in \cite{xtremespeech}) are shown below. For full definitions, we refer readers to the original paper.

\begin{enumerate}
	\item Derogatory extreme speech: ``Text that crosses the boundaries of civility within specific contexts and targets individuals/groups based on protected characteristics.''
	\item Exclusionary extreme speech: ``Expressions that call for or imply excluding historically disadvantaged and vulnerable groups based on protected attributes such as national origin, gender and sexual orientation.''
	\item Dangerous extreme speech: ``Text that has a reasonable chance to trigger harm against target groups.''
\end{enumerate}

\subsection{Few-Shot Learning}

With the general (relative) lack of non-English data in the domain of hate speech, as well as due to the difficulty of sourcing high-quality hate speech data, few-shot learning emerges as an attractive option for model development.

In few-shot learning settings, training shot selection is of great importance to model performance \cite{zheng-etal-2022-fewnlu,köksal2022meal}. This is especially salient in multilingual settings, where manual evaluation or prompt engineering might be challenging due to language barriers.

We propose the use of \dictionary to aid shot selection, allowing for more culturally-informed sampling of training examples. Instead of randomly selecting shots, we show how \dictionary can be used to select examples to cover a wider range of target groups and slurs in each cultural context.

We evaluate our proposed method using SetFit \cite{setfit}, training a multilingual SentenceBERT model\footnote{\url{https://huggingface.co/sentence-transformers/paraphrase-multilingual-mpnet-base-v2}} to discriminate between hateful and non-hateful speech.

\subsubsection{Data}

Training data comes from HASOC \cite{hasoc} and evaluation data comes from the Multilingual HateCheck benchmark \cite{rottger-etal-2022-multilingual}, on the German and Hindi subsets.  HASOC is a multilingual dataset of hate speech as sourced from Twitter. We focus on the binary classification task of HASOC, where tweets are classified as either hateful or neutral. The MHC benchmark is a suite with functional tests covering a wide range of hate speech categories.

To simulate a few-shot setting, we randomly sample 128 examples from HASOC German and Hindi each. This forms our total training set. We sample three sets for each languages with different seeds and report averaged results. We aim to investigate whether culturally-informed shot selection (via \dictionary) improves performance over random shot selection. We work with three dataset sizes: 32, 64 and 96.

\subsubsection{Shot Selection}

\textbf{Sampling Method Comparison.} For random selection, we sample shots without replacement. For the lexicon-based selection, we work in two steps: \begin{enumerate*}[label=(\roman*)]
	\item select all training examples that contain a slur or a target group term,
	\item further sample randomly to reach the desired training set size
\end{enumerate*}.
In \tabref{target_slur_dist} we show the distribution of slurs and targets in each sampled set. As expected, the randomly sampled sets do contain slurs and targets, although less frequently than in the \dictionary-sampled sets.

\begin{table}[]
	\centering
	\begin{tabular}{ccccc}
		\toprule
		\multirow{2}{*}{Set} & \multicolumn{2}{c}{German} & \multicolumn{2}{c}{Hindi} \\
		& S & T & S & T \\ \midrule
		All$_{128}$ & 12 & 10 & 12 & 9 \\
		Random$_{32}$ & 2 & 0 & 3 & 3 \\
		Random$_{64}$ & 3 & 4 & 5 & 4 \\
		Random$_{96}$ & 8 & 8 & 6 & 4 \\
		Lexicon$_{xx}$ & 12 & 10 & 12 & 9 \\ \bottomrule
	\end{tabular}
	\caption{Distribution of slurs (S) and targets (T) in German and Hindi sets.}
	\tablabel{target_slur_dist}
\end{table}

\textbf{Complementing Data.} Developers tackling hate speech online might try complementing their datasets with more data to improve performance. Since data collection and annotation is expensive and challenging, especially in low-resource languages, it is imperative that the collected data is of high quality.

To simulate this setting, we are procuring a few more training examples using \dictionary. On top of the sampled datasets as well as the entire dataset, we are further sampling from HASOC German and Hindi 16 training examples containing a target term and 16 examples containing a slur. These 32 examples are then added to the previous training sets and we repeat few-shot training (from scratch). For a fair comparison, we also sample 32 examples randomly and compare performance.

With this experiment we are aiming to investigate whether we can boost performance of a given training set by collecting training data specifically containing terms from \dictionary. Thus, developers in need of more data can query for terms found in \dictionary. While in our case we are merely sampling from HASOC, an already annotated hate speech dataset, this method could be used generally by querying for keywords on social media platforms and annotating as is practice.


\section{Interpretability through a Cultural Lens}
\seclabel{interpretability}

\subsection{\lime Analysis}

To showcase the usefulness of \dictionary in hate speech detection model interpretability, we analyze predictions (as reported by the authors) of \dataset.

As part of their study, \citet{xtremespeech} conduct an interpretability analysis of \texttt{mBERT} predictions for a three-way classification task to identify the extremity of text (derogatory, exclusionary or dangerous). Using \lime, they identify the top-10 words contributing the most to \texttt{mBERT}'s predictions (shown in \tabref{lime_explainability}).

In brief, the authors conclude that target group names (such as religious groups) and slurs contribute prominently to model predictions.
This exercise was performed in close collaboration with the annotators, who had to manually examine the identified top-contributing words. This process requires significant annotator effort and thus does not scale to practical settings.

\begin{table}[]
	\centering
	\small
	\renewcommand{\arraystretch}{0.91} 
	\begin{tabular}{cccc}
		\toprule
		Brazil & Germany & India & Kenya \\ \midrule
		fechar & Politiker & {\color{blue}muslims} & cows\\
		{\color{red}Ucranizar} & Grünen & {\color{blue}Muslim} & ruto \\
		{\color{red}ucranizar} & {\color{mypurple}{\scriptsize Mohammedaner}} & {\color{blue}muslim} & {\color{blue}luo} \\
		{\color{red}safada} & {\color{blue}Juden} & {\color{blue}Muslims} & {\color{red}wajinga} \\
		prender & Merkels & ko & {\color{blue}kikuyu} \\
		lixo & Merkel & {\color{red}mullo} & stupid \\
		coisa & Regierung & {\color{blue}Rohingyas} & idiot \\
		kkkkk & Opfer & da & looting \\
		{\color{red}Vagabundo} & {\color{blue}Islam} & {\color{red}suvar} & {\color{blue}tangatanga} \\
		{\color{red}traveco} & {\color{blue}Moslems} & dava & ujinga \\ \bottomrule
	\end{tabular}
	\caption{Top words contributing to \texttt{mBERT}'s
          predictions. Blue: {\color{blue}target group}. Red: {\color{red}slur}. Purple: {\color{mypurple}both}.}
	\tablabel{lime_explainability}
\end{table}

With \dictionary, we can automate the process, significantly reducing cost and time consumption.
We find that
in Brazil, there are 5 slurs; in Germany, 1 slur and 4 targets; in India, 2 slurs and 5 targets and in Kenya, 1 slur and 3 targets.

It is obvious that the model relies on the presence of slurs to make decisions, since slurs are predominantly used in hateful contexts. The model, though, also relies heavily on target group denotations when making predictions. Due to the (naturally) heightened presence of target groups in hate speech training data, models might learn to associate these otherwise innocuous terms with hate speech, overemphasizing their correlation with harmful content.
With \dictionary, we are able to identify this erroneous behavior of \texttt{mBERT} and potentially work on mitigating this effect.

\subsection{Change of LM Representation}

To investigate the effect training on slurs and target group names has on language models, we compare \texttt{mBERT}'s representation of lexicon terms before and after finetuning for India and Kenya. We finetune \texttt{mBERT} for the three-way classification task of \citet{xtremespeech} on the Indian and Kenyan sets.\footnote{Access was granted to use the Indian and Kenyan sets.} Specifically, we extract the representation of the 8th layer\footnote{The 8th layer has been found to contain useful representation in multilingual models \cite{jalili-sabet-etal-2020-simalign,dufter-schutze-2020-identifying}.} for the desired tokens\footnote{When a word spans more than one token, we average the representation of each token of the word.} and compute the cosine similarity with the corresponding representation in vanilla \texttt{mBERT}.

As a baseline, we compare the change of random words and stopwords from each country.
Random words were sampled from the development set of \citet{xtremespeech}, matching in number the \dictionary terms.
In \tabref{keywords_rep} we show that in Kenya, the representation of slurs changed the most after finetuning, with the representation of targets closely behind.
This indicates that vanilla \texttt{mBERT} has not adequately learned Kenyan slurs and target groups, since their representations changes significantly after we expose the model to the terms.
In India, on the other hand, the representation of slurs changed less than that of random words.
Considering the low performance of the Indian models (as reported by \citet{xtremespeech}) and the fact that targets make up half the list of top-contributing words (\tabref{lime_explainability}), we hypothesize the finetuned model has not sufficiently associated slurs with extreme speech.

\begin{table}[]
	\centering
	\begin{tabular}{ccc}
		\toprule
		& India & Kenya \\ \midrule
		Slurs & 1.85  & 2.54 \\
		Targets & 2.19 & 2.51 \\
		Stop & 2.16 & 2.20 \\
		Random & 2.13 & 2.38 \\ \bottomrule
	\end{tabular}
	\caption{Cosine similarity of representations between original \texttt{mBERT} and models finetuned on the Indian and Kenyan sets.}
	\tablabel{keywords_rep}
\end{table}

\section{Few-Shot Learning}

\subsection{Setup}

In our experiments, we are comparing three sets of training data: randomly-sampled (denoted with Random$_{xx}$), lexicon-based sampling (denoted with Lexicon$_{xx}$) and the entire training set (denoted with All$_{xx}$), where $xx$ denotes the training set size (by default equal to 128 for All$_{xx}$).

Further, we denote with $+l$ the sets complemented with 32 training examples additionally sampled using lexicon terms and we denote with $+r$ the sets complemented with 32 training examples additionally sampled randomly.

\subsection{Results}

\begin{figure*}
	\centering
	\begin{subfigure}{.5\textwidth}
		\centering
		\includegraphics[width=.9\textwidth]{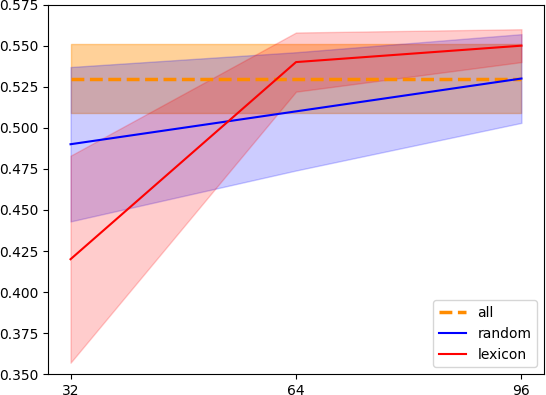}
		\caption{German}
	\end{subfigure}%
	\begin{subfigure}{.5\textwidth}
		\centering
		\includegraphics[width=.9\textwidth]{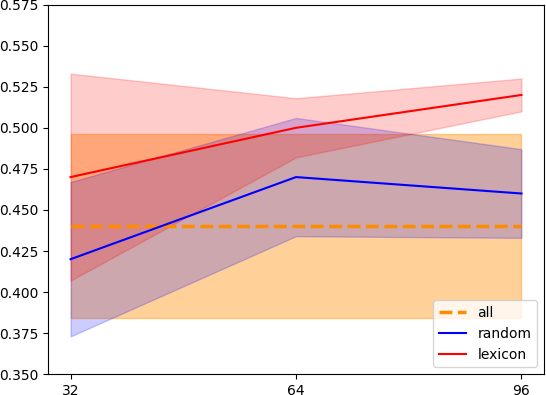}
		\caption{Hindi}
	\end{subfigure}
	\caption{Macro F1 (areas within one standard deviation are shaded) for MHC German (a) and Hindi (b). Lexicon-based sampling (red) outperforms both random sampling (blue) and All$_{128}$ (orange) for set sizes 64 and 96.}
	\figlabel{german_hindi_res}
\end{figure*}

\begin{table*}[ht]
	\begin{subtable}{0.5\textwidth}
		\centering
		\begin{tabular}{lcc}
			\toprule
			Set & F1 & $\Delta$ \\ \midrule
			Random$_{32+l}$ & $0.54_{3.2}$ & +0.05 \\
			Random$_{32+r}$ & $0.51_{3.6}$ & +0.02 \\
			Random$_{64+l}$ & $0.54_{2.1}$ & +0.03 \\
			Random$_{64+r}$ & $0.54_{1.9}$ & +0.03 \\
			Random$_{96+l}$ & $0.56_{1.2}$ & +0.03 \\
			Random$_{96+r}$ & $0.51_{0.9}$ & -0.02 \\ \midrule
			Lexicon$_{32+l}$ & $0.56_{1.2}$ & +0.14 \\
			Lexicon$_{32+r}$ & $0.51_{3.0}$ & +0.09 \\
			Lexicon$_{64+l}$ & $0.56_{0.9}$ & +0.02 \\
			Lexicon$_{64+r}$ & $0.55_{1.8}$ & +0.01 \\
			Lexicon$_{96+l}$ & $0.59_{0.4}$ & +0.04 \\
			Lexicon$_{96+r}$ & $0.56_{0.9}$ & +0.01 \\ \midrule 
			All$_{128+l}$ & $0.57_{1.1}$ & +0.04 \\
			All$_{128+r}$ & $0.54_{2.2}$ & +0.01 \\ \bottomrule
		\end{tabular}
		\caption{German}
	\end{subtable}
	\begin{subtable}{0.5\textwidth}
		\centering
		\begin{tabular}{lcc}
			\toprule
			Set & F1 & $\Delta$ \\ \midrule
			Random$_{32+l}$ & $0.47_{3.3}$ & +0.05 \\
			Random$_{32+r}$ & $0.44_{4.2}$ & +0.02 \\
			Random$_{64+l}$ & $0.50_{1.8}$ & +0.03 \\
			Random$_{64+r}$ & $0.48_{2.4}$ & +0.03 \\
			Random$_{96+l}$ & $0.51_{1.1}$ & +0.03 \\
			Random$_{96+r}$ & $0.48_{2.9}$ & -0.02 \\ \midrule
			Lexicon$_{32+l}$ & $0.50_{0.9}$ & +0.14 \\
			Lexicon$_{32+r}$ & $0.48_{2.0}$ & +0.09 \\
			Lexicon$_{64+l}$ & $0.52_{0.3}$ & +0.02 \\
			Lexicon$_{64+r}$ & $0.51_{0.9}$ & +0.01 \\
			Lexicon$_{96+l}$ & $0.55_{0.3}$ & +0.04 \\
			Lexicon$_{96+r}$ & $0.51_{0.6}$ & +0.01 \\ \midrule
			All$_{128+l}$ & $0.49_{1.3}$ & +0.04 \\
			All$_{128+r}$ & $0.49_{1.0}$ & +0.01 \\ \bottomrule
		\end{tabular}
		\caption{Hindi}
	\end{subtable}
	\caption{Macro F1 (standard deviation as subscript) and difference with the non-complemented baseline ($\Delta$), for MHC German (a) and Hindi (b).}
	\tablabel{complement_res}
\end{table*}

\textbf{Main Results.} In \figref{german_hindi_res}, we compare macro F1-scores between \dictionary- and randomly-sampled training sets as well as the set containing all available training examples (All$_{128}$).

In German, excluding the sets with a size of 32 which perform poorly, training sets sampled via \dictionary outperform the corresponding randomly-sampled sets. Both Lexicon$_{64}$ and Lexicon$_{96}$ outperform all randomly-sampled sets, as well as All$_{128}$. At the same time, with the lexicon-based sampling method, performance is more consistent across runs, especially at sizes 64 and 96 which have a small standard deviation.

The difference between lexicon- and random-based sampling is starker in the Hindi set. Lexicon-based training sets outperform all other baselines and, like in the German experiments, standard deviation is minimized when sampling with \dictionary, providing better stability.

It is noteworthy that \dictionary-based training sets regularly outperform All$_{128}$. We hypothesize this is due to a higher concentration of high-quality training data in Lexicon$_{xx}$ sets. In few-shot settings, the importance of each training example is magnified and thus noise can potentially affect performance disproportionately \cite{zheng-etal-2022-fewnlu,mosbach2021on}.
This is in-line with other works: for example, in \citet{schick-schutze-2021-exploiting}, results on AGNews are worse when using the largest training set. In Section \ref{predicting_shots}, we are further investigating this phenomenon.

\textbf{Data Complementing Results.} In \tabref{complement_res}, we show results of our data complementing experiments.
In these experiments, we are adding to the training data 32 examples sampled either randomly ($+r$ in notation) or via \dictionary ($+l$ in notation). We see a consistent increase in performance when adding examples sampled with our proposed method.
Namely, Lexicon$_{xx+l}$ sets consistently perform better than
both the original (without data complementing) baselines and the $+r$ variants.
When complementing the Random$_{xx}$ sets, performance is slightly more inconsistent, although the $+l$ sets still perform better than the $+r$ variants.
Further, Lexicon$_{xx+l}$ sets have a low standard deviation, while Random$_{xx+l}$ and Random$_{xx+r}$ have a consistently higher standard deviation, showing that lexicon-based sampling is overall more consistent. In general, wherever we complement using \dictionary ($+l$ sets), performance is better compared to both the original and the randomly-complemented ($+r$) sets.

\subsection{Ablation Study - Predicting Shots}
\label{predicting_shots}

A reason why our lexicon-based sampling method works better than random sampling may be that it retrieves less noise and fewer ambiguous examples. It has been previously shown that hard-to-learn examples (such as text that is ambiguous, misannotated or difficult to predict) do not contribute positively to model development \cite{swayamdipta-etal-2020-dataset}. We hypothesize that our sampling method replaces a large portion of these low-quality examples with high-quality, information-rich examples.

To investigate whether our hypothesis holds true, we develop a hate speech model and \textit{apply it on our examined training sets}. Since lexicon-sampled training sets are bound to contain more informative and unambiguous examples, they should be easier to classify correctly than randomly-sampled examples.

For this ablation study, we finetune an \texttt{XLM-RoBERTa-base} model \cite{xlmr} separately on all the originally available German and Hindi data from HASOC \cite{hasoc}, excluding the 128 training examples sampled for our experiments, for a total of 2245 training examples for German and 2835 for Hindi. Then, we apply the two resulting models on our few-shot learning training sets.\footnote{We only predict shots from a randomly-chosen training set for each language instead of all three used previously.}

We show (\tabref{pred_shots}) that examples sampled with our \dictionary are easier to classify correctly.
In Hindi, the lexicon-based set is easier by 0.02 over the other sets. In German, performance on the lexicon-based set is 0.05 higher than the randomly sampled set and 0.10 higher than the entire set. Thus, we can infer that with our lexicon-based sampling method, easier examples are sourced more often than harder-to-classify ones.

\begin{table}[]
	\centering
	\begin{tabular}{ccc}
		\toprule
		& Germany & India \\ \midrule
		Lexicon & 0.61 & 0.55 \\
		Random & 0.56 & 0.53 \\
		All & 0.51 & 0.53 \\ \bottomrule
	\end{tabular}
	\caption{F1-score of classifying training shots.}
	\tablabel{pred_shots}
\end{table}

Manual inspection of prediction errors
of examples
contained
in the total training set
but not in the lexicon-based set shows a high rate of low-quality text and noise (\tabref{manual_inspec}).
Example 0 has been annotated as hateful even though it is just noise, containing only an account mention. Example 1 is an ambiguous (given the lack of context) example containing a sarcastic comment against a political party (die Grünen / Greens), classified by the model as hateful. Example 2 is also ambiguous, containing sarcasm against a right-wing party in Germany (AfD). All ambiguous examples (1-3) are short tweets that mention political entities.\footnote{While political entities are an integral part of society, they are not target groups and were not added to our lexicon.} In a politically charged environment, these short texts do not provide enough context for the model to adequately learn whether the example is hateful or not. Therefore, adding these examples in our training set is not beneficial.

\begin{table}[]
	\centering
	\small
	\begin{tabular}{ cp{4cm}c  }
		\toprule
		ID & Text & Type \\ \midrule
		0 & @Hartes\_Geld & Noise \\
		1 & Ja so tierlieb sind die grünen & Ambiguous \\
		2 & @SaschaUlbrich @Mundaufmachen @AfD super, gut gemacht! auf jeden Fall ``retweeten"! & Ambiguous \\
		3 & Wer soll jetzt die SPD führen? & Low-content \\ \bottomrule
	\end{tabular}
	\caption{Manual inspection of model prediction errors.}
	\tablabel{manual_inspec}
\end{table}

\section{Conclusion}
\label{sec:conclusion}

In our work, we curate \dictionary, a lexicon of slurs and targets of hate speech for the countries of Brazil, Germany, India and Kenya, with the goal of improving model development.

With our lexicon, we show how models rely on slurs and target group denotations when making predictions in hate speech tasks. The over-reliance on target group names may lead to further marginalization of targets of hate speech, with models flagging as hateful innocuous text containing these terms. With \dictionary, this erroneous behavior is unveiled and researchers can focus on mitigating this bias.

We also demonstrate how \dictionary can be used for few-shot learning. We evaluate on the German and Hindi subsets of the Multilingual HateCheck benchmark \cite{rottger-etal-2022-multilingual} and show that selecting training shots with a culturally-informed process (e.g., our lexicon of slurs and targets) can aid the development of hate speech classifiers. Namely, training sets sampled using \dictionary perform better than training sets sampled at random.

More abstractly, we provide evidence that curating sociocultural knowledge bases (e.g., in the form of lexicons) is pivotal in developing hate speech detection models. Sociocultural information is vital in understanding hate speech in given contexts,
and without it we risk developing models detached from the reality and experiences of the most vulnerable.
Thus, we advocate for a greater focus on bridging the knowledge gap between researchers and affected communities for the development of models better geared towards protecting target groups.

\section{Acknowledgement}

This work was funded by the European Research Council (grant \#740516).

\section{Ethical Considerations and Limitations}

\subsection{Ethics Statement}

In our work we are dealing with sensitive content in the form of hate speech against marginalized communities. We are not advocating for hate speech, but instead propose methods to aid in filtering out harmful content from online spheres and analyzing detection models with our proposed lexicon.

The lexicon was developed in collaboration with annotators familiar with sociocultural balances in their countries and communities, with the goal of creating a dictionary of terms useful for hate speech model development. A potential concern with a dictionary of hateful terms is that the terms will be publicized and could be subsequently used by hate speech peddlers to cause further harm. Since these terms were recorded specifically because they are already used extensively, the risk of additional harm from publicizing these terms is minimal. Moreover, in \dictionary we are collecting denotations of target groups. These may be based on ethnicity, religion, sexuality or other protected attributes. A potential concern is that we will be exposing the mentioned target groups. We argue that better understanding the harms faced by these communities outweighs the negatives and will provide more net-positive in the long term, while at the same time these groups were recorded due to the increased quantity of hateful content they receive.

\subsection{Limitations}

The lists of slurs and target groups in \dictionary are not exhaustive. While we took care to expand \dictionary as thoroughly as possible, we are limited by time and resources and could only cover a partial set of terms used online in relation to hate speech in the examined countries.

Further, the list of countries chosen is small: Brazil, Germany, India and Kenya. Ideally we would have included more countries and languages. More work needs to be done to expand this list.

\nocite{icons}
\bibliography{anthology}
\bibliographystyle{acl_natbib}

\clearpage
\appendix

\eat{
	\begin{table*}[ht]
		\begin{subtable}{0.5\textwidth}
			\centering
			\begin{tabular}{lccc}
				\toprule
				Set & F1 & P & R \\ \midrule
				Random$_{32}$ & $0.49_{4.7}$ & $0.52$ & $0.52$ \\
				Random$_{64}$ & $0.51_{3.6}$ & $0.53$ & $0.53$ \\
				Random$_{96}$ & $0.53_{2.7}$ & $0.55$ & $0.54$ \\ \midrule
				Lexicon$_{32}$ & $0.42_{6.3}$ & $0.48$ & $0.48$ \\
				Lexicon$_{64}$ & $0.54_{1.8}$ & $0.54$ & $0.54$ \\
				Lexicon$_{96}$ & $0.55_{1.0}$ & $0.55$ & $0.56$ \\ \midrule
				All$_{128}$ & $0.53_{2.1}$ & $0.57$ & $0.54$ \\ \bottomrule
			\end{tabular}
			\caption{German}
		\end{subtable}
		\begin{subtable}{0.5\textwidth}
			\centering
			\begin{tabular}{lccc}
				\toprule
				Set & F1 & P & R \\ \midrule
				Random$_{32}$ & $0.42_{6.1}$ & $0.53$ & $0.52$  \\
				Random$_{64}$ & $0.47_{2.5}$ & $0.52$ & $0.51$  \\
				Random$_{96}$ & $0.46_{3.2}$ & $0.54$ & $0.51$  \\ \midrule
				Lexicon$_{32}$ & $0.47_{5.4}$ & $0.55$ & $0.52$ \\
				Lexicon$_{64}$ & $0.50_{1.1}$ & $0.54$ & $0.52$ \\
				Lexicon$_{96}$ & $0.52_{0.4}$ & $0.53$ & $0.53$ \\ \midrule
				All$_{128}$ & $0.44_{5.6}$ & $0.52$ & $0.51$ \\ \bottomrule
			\end{tabular}
			\caption{Hindi}
		\end{subtable}
		\caption{Macro F1 (standard deviation in subscript), Precision (P) and Recall (R) for MHC German (a) and Hindi (b).}
		\tablabel{res_table}
	\end{table*}
}

\end{document}